\documentclass[10pt,twocolumn,letterpaper]{article}

\usepackage[pagenumbers]{cvpr} 

\usepackage{graphicx}
\usepackage{amsmath}
\usepackage{amssymb}
\usepackage{booktabs}
\usepackage{multirow}

\usepackage[pagebackref,breaklinks,colorlinks]{hyperref}

\usepackage[capitalize]{cleveref}
\crefname{section}{Sec.}{Secs.}
\Crefname{section}{Section}{Sections}
\Crefname{table}{Table}{Tables}
\crefname{table}{Tab.}{Tabs.}
\usepackage{color}

\def\cvprPaperID{8643} 
\def\confName{CVPR}
\def\confYear{2022}

\begin{document}

\title{ESGN: Efficient Stereo Geometry Network for Fast  3D Object Detection}
\author{Aqi Gao, Yanwei Pang, Jing Nie, Jiale Cao and Yishun Guo \\
Tianjin University\\
{\tt\small {gaoaqi,pyw,jingnie,connor,guoyishun}@tju.edu.cn}}

\maketitle


\begin{abstract}
Fast stereo based 3D object detectors have made great progress recently. However, they lag far behind  high-precision stereo based methods in accuracy. We argue that the main reason is due to the poor geometry-aware feature representation in 3D space. To solve this problem, we propose an efficient stereo geometry network (ESGN). The key in our ESGN is an efficient geometry-aware feature generation (EGFG) module. Our EGFG module first uses a stereo correlation and reprojection module to construct multi-scale stereo volumes in camera frustum space, second employs a multi-scale BEV projection and fusion module  to generate multiple geometry-aware features. In these two steps, we adopt deep multi-scale information fusion  for discriminative geometry-aware feature generation, without any 
complex aggregation networks.  In addition, we introduce a  deep geometry-aware feature distillation scheme to guide stereo  feature learning with a LiDAR-based detector. The experiments are performed on the classical KITTI dataset. On KITTI test set, our ESGN outperforms the fast state-of-art-art detector YOLOStereo3D by 5.14\% on mAP$_{3d}$ at 62$ms$. To the best of our knowledge, our ESGN achieves a best trade-off between accuracy and speed. We hope that our efficient stereo geometry network can
provide more possible directions for fast 3D object detection. Our source code will be released.
\end{abstract}


\maketitle

\section{Introduction}

3D object detection is an important but challenging computer vision task, which is essential for automatic driving. Though LiDAR-based 3D object detection approaches \cite{ Qi_PointNet_CVPR_2017,2019PointRCNN,Qi_PointNet++_NIPS_2017} have high accuracy, they suffer from the expensive hardaware cost and low resolution. Compared with LiDAR-based 3D object detection approaches, stereo-based 3D object detection approaches \cite{Li_StereoRCNN_CVPR_2019,Peng_IDA3D_CVPR_2020,wang_PseudoLiDAR_2019} adopt the low-cost optical camera and can provide dense 3D information. In this paper, we focus on stereo-based 3D object detection. The stereo-based 3D object detection approaches can be mainly divided into camera frustum space based methods, pseudo LiDAR based methods, and voxel based methods.


\begin{figure}[t]
\centering
\includegraphics[width=0.999\linewidth]{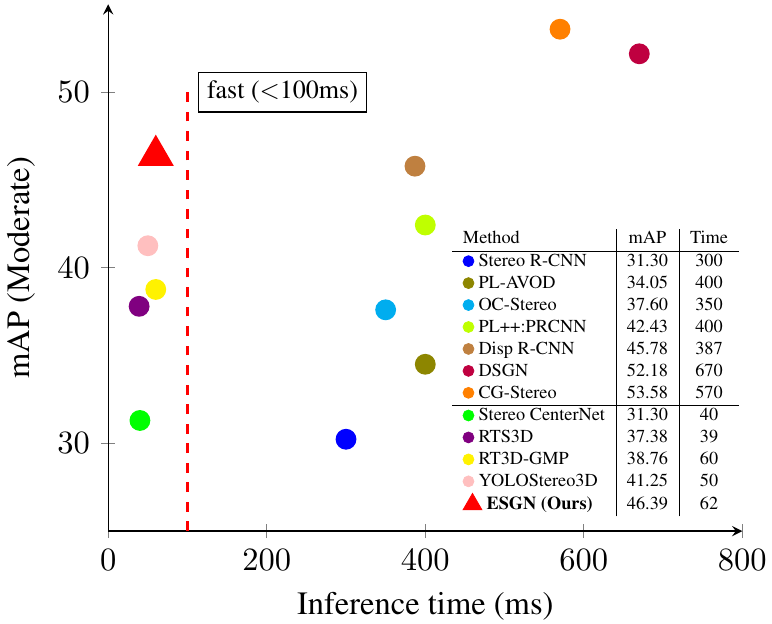}
\caption{Accuracy (mAP) and inference time (ms) comparison of some state-of-the-art stereo 3D object detection methods on KITTI  test (moderate) set \cite{kitti}. The inference time of most methods, except YOLOStereo3D \cite{liu2021yolostereo3d}, are taken from the official KITTI leaderboards. For a fair comparison, the inference time of  YOLOStereo3D and our ESGN are reported on a single NVIDIA RTX3090. Our ESGN achieves a best trade-off between accuracy and speed.}
   \label{fig:trade-off}
\end{figure}

\begin{figure*}
\begin{center}
\includegraphics[width=0.99\linewidth]{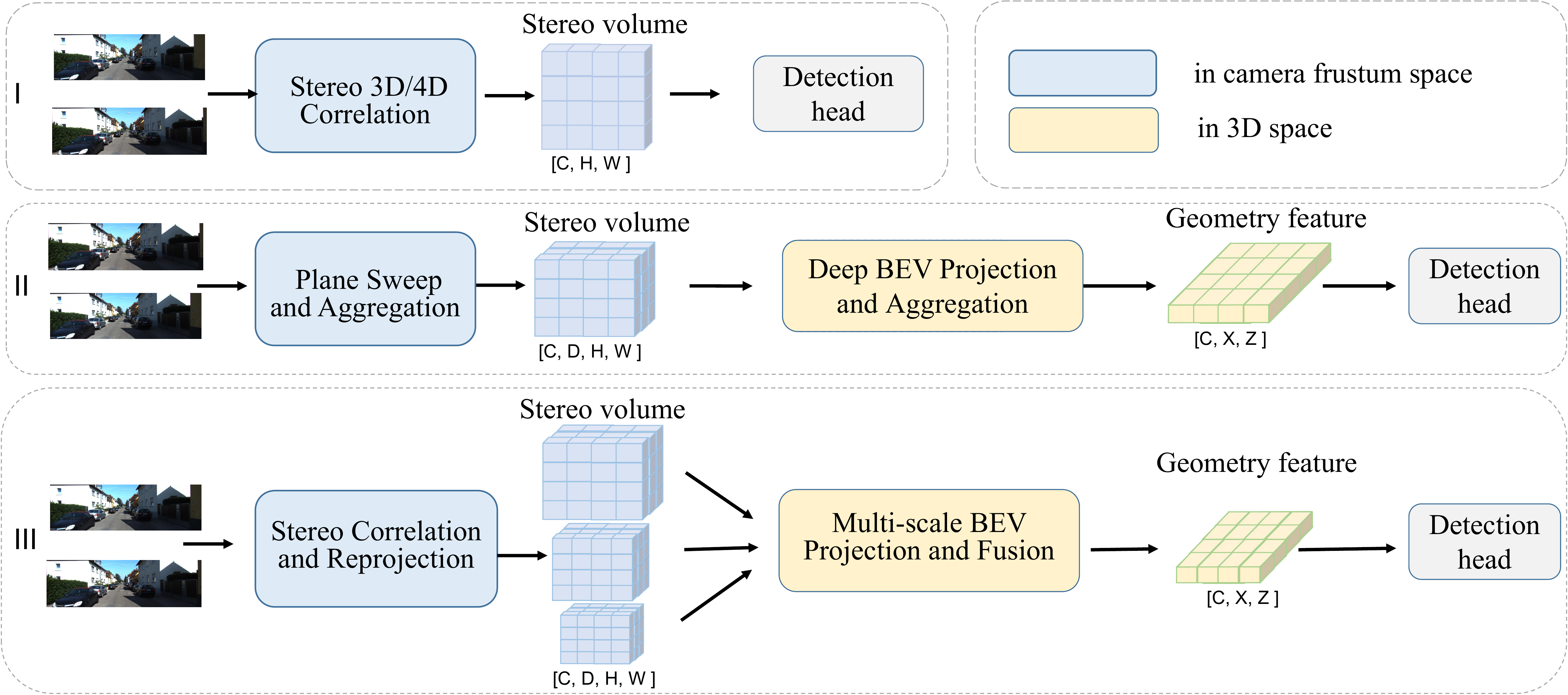}
\caption{Structure comparison of different stereo methods. (I) Camera frustum space based method (\textit{e.g.,} YOLOStereo3D \cite{liu2021yolostereo3d}). It uses a stereo 3D/4D correlation module to generate stereo volume in camera frustum space for 3D detection, which faces the issue of object distortion. (II) Voxel based method (\textit{e.g.,} DSGN \cite{chen2020dsgn}). In the steps of stereo volume and geometry-aware feature generations, it adopts the heavy 3D and 2D aggregation network, resulting in a slow speed. (III) Our efficient stereo geometry netork (ESGN). Compared to 
voxel based method DSGN, our ESGN adopts deep multi-scale information fusion, instead of heavy 3D and 2D aggregation networks, for geometry-aware feature generation.}
\label{fig:simple-module}
\end{center}
\end{figure*}

As shown in Fig. \ref{fig:trade-off}, we compare the accuracy and inference time of some state-of-the-art stereo 3D object detection methods, where  we call the methods with the inference time less than 100$ms$ as fast stereo 3D object detection methods. Most of these fast methods belong to camera frustum space based methods (\textit{e.g.,} YOLOStereo3D  \cite{liu2021yolostereo3d}) and pseudo LiDAR based methods (\textit{e.g.,} RT3D-GMP \cite{RT3D-GMP} and RT3DStereo \cite{RT3DStereo}).
The camera frustum space based methods \cite{liu2021yolostereo3d} (see Fig. \ref{fig:simple-module}(I)) first employ a stereo 3D/4D correlation module to generate stereo volume in camera frustum space and second perform 3D object detection directly on stereo volume.
Pseudo LiDAR based methods first obtain the depth map by a fast depth estimation network and transform it into point cloud. After that, they employ a light-weighted LiDAR-based detector to detect objects. These fast stereo 3D object detection approaches achieve fast speed by adopting the light-weighted modules in each processing step. However, camera frustum space based methods lack feature representation in 3D space, resulting in the issue of object distortion, while pseudo LiDAR based methods are sensitive to the error of fast depth estimation network.



Compared with these fast stereo 3D methods, voxel based methods \cite{chen2020dsgn,guo2021liga} are dominant in accuracy. Fig. \ref{fig:simple-module}(II) shows the basic pipeline of voxel based methods. They first use a plane sweep and aggregation module to generate stereo volume in camera frustum space.
After that, they apply a deep BEV projection and aggregation module to extract geometry-aware feature in 3D space and perform 3D detection on geometry-aware feature. In these two steps, the voxel based methods employ heavy 3D and 2D aggregation networks to extract discriminative features.  Due to these heavy operations, voxel based methods are very slow in speed.


We argue that, compared with voxel based methods, the issue that  impedes the performance of  fast stereo 3D methods is the  poor geometry-aware feature representation in 3D space. To solve this issue, we propose an efficient stereo geometry network (ESGN) in  Fig. \ref{fig:simple-module}(III). The key is an efficient 3D geometry-aware feature generation module
(EGFG), which consists of a stereo correlation  and reprojection (SCR) module and a multi-scale BEV projection and fusion (MPF) module. 
We first use the SCR module to generate multi-scale stereo volumes in camera frustum space.
After that, we apply the MPF module to convert multi-scale stereo volumes  into multiple geometry-aware features in 3D space. Finally, we perform 3D detection on one of  geometry-aware features. 
In addition, we introduce a deep geometry-aware feature distillation scheme to guide feature learning, where a LiDAR-based detector is designed to provide deep supervisions in multiple levels. 
Compared with fast stereo method YOLOStereo3D \cite{liu2021yolostereo3d}, our ESGN generates geometry-aware feature in 3D space and avoids object distortion in camera frustum space. Compared with voxel based method DSGN \cite{chen2020dsgn}, our ESGN adopt efficient deep multi-scale information fusion and has a much faster speed. 
We perform the experiments on the classical KITTI dataset \cite{kitti}. As shown in Fig. \ref{fig:trade-off}, our proposed ESGN achieves an optimal trade-off between accuracy and speed. Our contributions can be summarized as follows:
\begin{itemize}
\item We propose an efficient stereo geometry-aware feature network (ESGN) for fast 3D object detection. The key module is an efficient  geometry-aware feature generation (EGFG) module. EGFG extracts discriminative geometry-aware features in 3D space by adopting deep multi-scale information fusion in the steps of both stereo volume and geometry-aware feature generations.
\item We further introduce a deep geometry-aware feature distillation (DGFD) scheme. DGFD uses a LiDAR-based detector to extract multi-level geometry-aware features and then employs these features to guide stereo feature learning. 
\item We perform the experiments on the classical KITTI dataset \cite{kitti}. On the moderate test set, our proposed ESGN achieves an $AP_{3d}$ of 46.39\% at a speed of 62$ms$, obtaining an optimal trade-off between accuracy and speed. Compared to fast YOLOStereo3D \cite{liu2021yolostereo3d}, our ESGN provides an absolute gain of 5.14\% at a comparable speed. 
\end{itemize}




\section{Related Work}
\label{sec:formatting}
Compared to 2D object detection \cite{Ren_FasterRCNN_NIPS_2015,Tian_FCOS_ICCV_2019,Duan_CenterNet_ICCV_2019}, 3D object detection aims to classify and localize objects in 3D space, which is more challenging and very useful for real applications. In this section, we first introduce the related works of stereo 3D object detection. After that, we give a short review on the related LiDAR-based 3D object detection and knowledge distillation.

\subsection{Stereo 3D Object Detection}
As mentioned earlier, stereo 3D object detection can be mainly divided into three classes: camera frustum space based methods, pseudo LiDAR based methods, voxel based methods. Camera frustum space based methods extract the features in image coordinate system for 3D object detection.
Stereo RCNN \cite{Li_StereoRCNN_CVPR_2019} first predicts a rough 3D bounding box based on the combined RoI features from the left and right images, and second conducts a bundle adjustment optimization for final 3D bounding box prediction. IDA-3D \cite{Peng_IDA3D_CVPR_2020} builds a cost volume from left and right RoI features to generate the  center point depth for 3D object detection. 

Pseudo LiDAR based methods explore to convert stereo 3D detection into LiDAR-based 3D detection. Pseudo-LiDAR \cite{wang_PseudoLiDAR_2019} is one of the earliest pseudo LiDAR based methods. It transforms the depth map of stereo images into the point cloud data and performs 3D point cloud detection. Pseudo-LiDAR++ \cite{you_Pseudo-LiDAR++_2020} introduces a depth cost volume to  generate depth map directly. OC-Stereo \cite{Pon_ocstereo_2020} and Disp RCNN \cite{sun_disprcnn_2020} only consider the foreground regions of point cloud and achieve a better performance. ZoomNet \cite{xu_Zoomnet_2020} improves the disparity estimation by enlarging the target region.

Voxel based methods extract voxel features in 3D space to detect objects. DSGN \cite{chen2020dsgn} coverts stereo volume in camera frustum space into the volume in 3D space space volume to better represent the object 3D structure. LIGA \cite{guo2021liga} applies a LiDAR-based detector detector  as a teacher model to guide geometry-aware feature learning. With discriminative geometry-aware features, these methods are dominate in accuracy. However, due to heavy 3D convolutions and heavy 3D/2D aggregation networks, these methods are insufficient enough to be applied in practice. 
\begin{figure*}
\begin{center}
\includegraphics[width=0.98\linewidth]{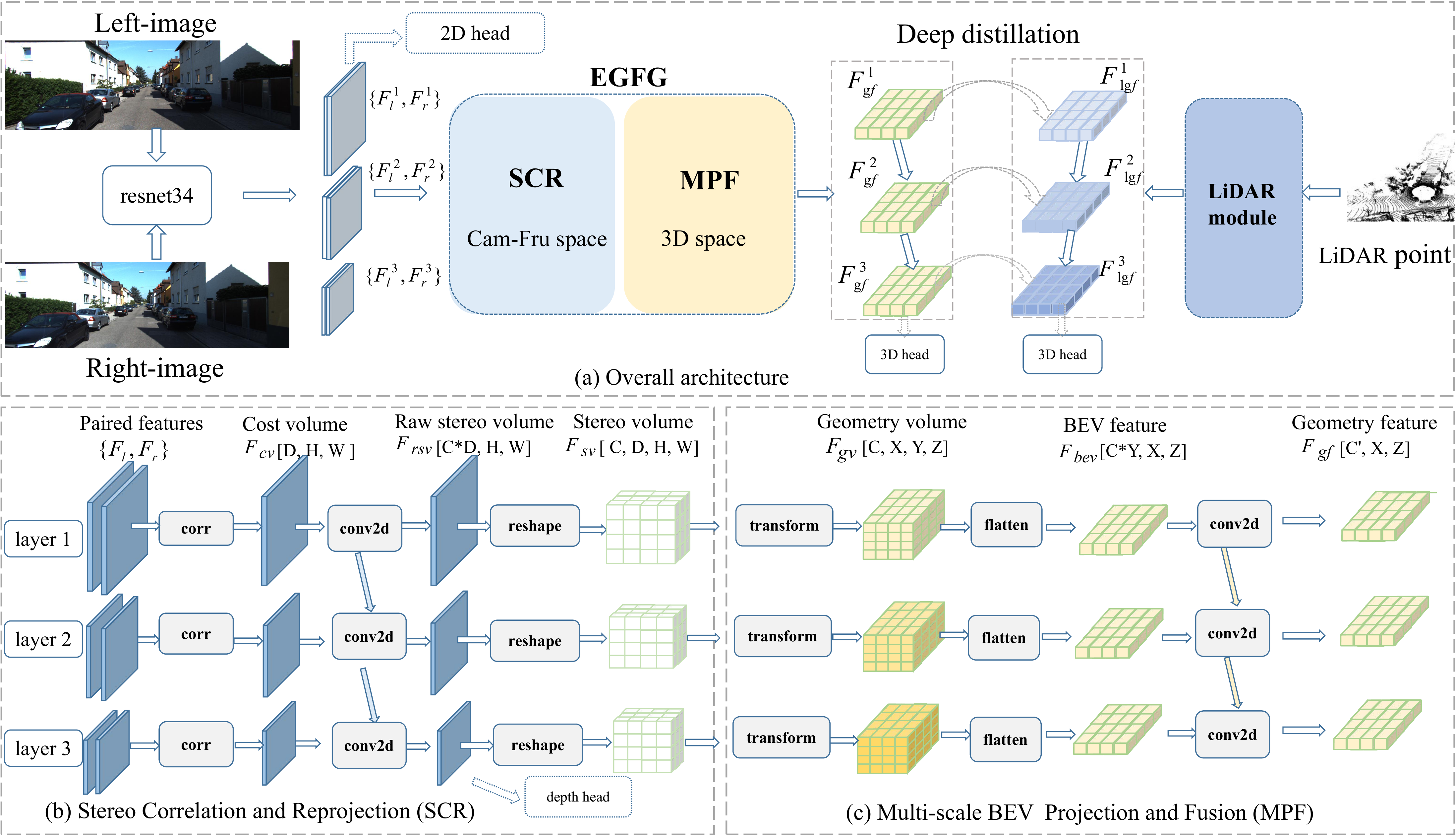}
\caption{(a) Overall architecture of our ESGN. Given a paired of input images, we adopt the efficient ResNet-34 \cite{He_ResNet_CVPR_2016} to extract multi-scale paired feature maps ($\{F_l^1,F_r^1\}, i=1,2,3$). Then, we employ our proposed efficient geometry-ware feature generation (EGFG) module to generate multiple geometry-ware features ($F_{gf}^i, i=1,2,3$). In addition, we introduce a LiDAR-based detector for deep feature distillation. (b) Stereo correlation and reprojection (SCR) module converts multi-scale paired feature maps to multiple stereo volumes in camera frustum space. (c) Multi-scale BEV projection and fusion (MPF) module converts camera frustum space to 3D space for geometry-aware feature generation.}
\label{fig:arch}
\end{center}
\end{figure*}
Fast stereo 3D objection detection approaches mainly belong to camera frustum space based and pseudo LiDAR based methods. Pseudo LiDAR based methods RT3D-GMP \cite{RT3D-GMP} and RT3DStereo \cite{RT3DStereo} use a light-weighted depth estimation module to generate depth map. After that, they transform the depth map into the point cloud data and use a light-weighted 3D point cloud detector to perform 3D detection. With these light-weighted modules, these methods have a high inference speed. 
Camera frustum space based method Stereo-Centernet \cite{shi2021sc} changes the anchor-based network in Stereo-RCNN \cite{Li_StereoRCNN_CVPR_2019} into a key-point based network for fast 3D prediction. YOLOStereo3D \cite{liu2021yolostereo3d} first adopts a stereo 3D/4D correlation module to generate  stereo volume in camera frustum space and second performs 3D object detection directly on the stereo volume. YOLOStereo3D achieves best performance among these fast methods. However, due to the poor 3D geometry-aware feature representation, YOLOStereo3D still lags far behind voxel based methods \cite{chen2020dsgn}. 
Our proposed methods aims to bring this gap by efficiently extracting discriminative geometry-aware feature and thus achieves a best trade-off between accuracy and speed. 

\subsection{LiDAR-based 3D Object Detection}
Compared to stereo 3D objection, LiDAR-based 3D object detection has a higher accuracy, but suffers from the expensive handaware cost and low resolution. The methods of LiDAR-based 3D object detection can be mainly divided into three parts: voxel based methods, point based methods, point-voxel based methods. Voxel based methods \cite{ZhouVoxelnet, AlexPointpillars,shipart2,YeHvnet} transform the irregular point clouds to the volumetric representations in compact shape and then extract the voxel features for 3D detection. Some backbones (\textit{e.g.,} PointNet \cite{Qi_PointNet_CVPR_2017} and PointNet++ \cite{Qi_PointNet++_NIPS_2017}) are proposed to directly extract the features from  irregular point cloud data.  Based on these backbones, point based methods  \cite{2019PointRCNN,yang3dssd,shipoint-gnn,Qi_voteNet_ICCV_2019, Xie_MLCVNet_CVPR_2020} directly perform 3D detection. 
Voxel based methods are usually efficient but faces the issue of information loss, while point based methods have easily achieve
larger receptive field but are inefficient. 
Point-voxel based methods \cite{shiPV-RCNN,chenfastpoint-rcnn,yangstd,Hesas3d} aim to integrate the advantages of voxel based methods and point based methods.

\subsection{Knowledge Distillation}
Knowledge distillation is first proposed  for network compression \cite{HintonDistilling}, where
a large and high-performance teacher network provides softened labels to  supervise the feature learning of a small student network. As a result, the student network can learn  better features with a small number of network parameters.
After that, some methods \cite{AdrianaFitnets,Byeonghocomprehensive} explore to make use of the knowledge from the intermediate layers of teacher network.
Recently, knowledge distillation has been successfully applied to
stereo 3D object detection \cite{guo2021liga}, which demonstrates that it is effective to use LiDAR-based detector  to guide stereo feature learning. We argue that the single-level distillation in \cite{guo2021liga} can not provide deep supervision for feature learning. To solve this problem, we propose a deep geometry-aware feature distillation scheme that provides deep multi-level supervisions on geometry-ware feature learning.


\section{Our Method}



In this section, we provide a detailed introduction about our efficient stereo geometry network (ESGN) for 3D object detection. Fig.~\ref{fig:arch}(a) shows the overall architecture of our proposed ESGN. We first employ an efficient deep model ResNet-34 \cite{He_ResNet_CVPR_2016} to extract multi-scale paired feature maps (\{$F_l^i$,$F_r^i$\}, $i=1,2,3$) from stereo input images. Based on these paired feature maps, we introduce a novel efficient geometry-aware feature generation (EGFG) module to generate multiple geometry-aware features ($F_{gf}^i$, $i=1,2,3$) with deep multi-scale information fusion. Our EGFG module contains a stereo correlation and reprojection (SCR) module and a multi-scale BEV preservation and fusion (MPF) module. To further enhance geometry-aware feature representation, we  introduce a deep geometry-aware feature distillation  (DGFD) scheme, where a LiDAR based detector is designed to extract multi-scale discriminative geometry-aware features ($F_{lgf}^i$, $i=1,2,3$) from point cloud data and then guide stereo geometry-aware feature learning. Finally, the geometry-aware feature $F_{gf}^3$ is fed to 3D prediction head for 3D object detection. 

Here, we first introduce the key efficient geometry-aware feature generation (EGFG) module in Sec.~\ref{sec:egfg}. After that, we describe the deep geometry-aware feature distillation (DGFD) scheme in  Sec.~\ref{sec:dgfd}. 
Finally, we describe the details of prediction heads, including 3D detection head and auxiliary heads, in Sec.~\ref{sec:head}.

\subsection{Efficient Geometry-Aware Feature Generation (EGFG)}
\label{sec:egfg}
The efficient geometry-aware feature generation (EGFG) module aims to convert multi-scale paired feature maps (\{$F_l^i$,$F_r^i$\}, $i=1,2,3$) extracted from stereo images to multiple geometry-aware features ($F_{gf}^i$, $i=1,2,3$) with deep multi-level fusion. The EGFG module consists of two sequential modules: a stereo correlation and reprojection (SCR) module and a multi-scale BEV projection and fusion (MPF)  module.  The SCR module first  generates multi-scale stereo volumes in camera frustum space with simple correlation and repojection operations. Then, the MPF  module converts multi-scale stereo volumes into 3D and BEV spaces to generate multiple geometry-aware features. 

\subsubsection{Stereo Correlation and Reprojection (SCR)}
\label{sec:scr}
As shown in Fig. \ref{fig:arch}(b), the SCR module takes multi-scale paired features (\{$F_l^i$,$F_r^i$\}, $i=1,2,3$) as inputs. With these paired features, the SCR module first extracts multi-scale cost volumes ($F_{cv}^i$, $i=1,2,3$) with stereo correlation operation and second generates multi-scale stereo volumes (\textit{i.e.,} $F_{sv}^i$, $i=1,2,3$) in camera frustum space with reprojection.

For a paired feature map \{$F_l^i$,$F_r^i$\}, the cost volume ($F_{cv}^i$) is first generated with stereo correlation as follows.
\begin{equation}
\label{eq:cost volume}
\begin{array}{c}
F_{cv}^i(d, h, w) =\frac{1}{C}\displaystyle\sum_{c=1}^{C}{F_l^i(c, h, w-d) * F_r^i(c, h, w+d)},
\end{array}
\end{equation}
where $C$ is the number of feature channel, and $d,h,w$ represent the indexes of depth, height, and width dimensions, respectively. If $w-d$ or $w+d$ are out of range of $F_l^i$ and $F_r^i$, we set the corresponding feature as zero. 

After that, we convert multi-scale  cost volumes ($F_{cv}^i$, $i=1,2,3$) to  multi-scale stereo volumes ($F_{sv}^i, i=1,2,3$) with our repojection operation, including multi-scale cost volume fusion and dimension reshaping. Assuming that the size of $F_{cv}^1$ is $D\times H \times W$, we first use a 2D convolution to generate raw stereo volume $F_{rsv}^1 \in \mathbb{R}^{(C*D)\times H \times W}$ and second reshape $F_{rsv}^1$ to stereo volume $F_{sv}^1\in \mathbb{R}^{C\times D\times H \times W}$. At the same time, we downsample raw stereo volume $F_{rsv}^1$ twice time and concatenate it with $F_{cv}^2$, which are fed to a 2D convolution to generate raw stereo volume $F_{rsv}^2 \in \mathbb{R}^{(C*D)\times H/2 \times W/2}$. The $F_{rsv}^2$ is reshaped to generate stereo volume $F_{sv}^2 \in \mathbb{R}^{C\times D\times H/2 \times W/2}$. Similarly, we generate stereo volume $F_{sv}^3$ with the inputs of $F_{rsv}^2$ and $F_{cv}^3$. The above steps of our reprojection operation can be summarized as the following equations. 
\begin{equation}
\left\{
\begin{array}{l}
F_{sv}^1 = f_{re}(F_{rsv}^1), F_{rsv}^1 = f_{conv}(F_{cv}^1),\\
F_{sv}^2 = f_{re}(F_{rsv}^2), F_{rsv}^2 = f_{conv}(f_{cat}(f_{avg}(F_{rsv}^1), F_{cv}^2)),\\
F_{sv}^3 = f_{re}(F_{rsv}^3), F_{rsv}^3= f_{conv}(f_{cat}(f_{avg}(F_{rsv}^2), F_{cv}^3)),\\
\end{array}
\right.
\label{eq:SV}
\end{equation}
where  $f_{re}$ represents dimension reshaping, $f_{conv}$ represents 2D convolution, $f_{cat}$ represents channel concatenation, $f_{avg}$ represents the downsampling operation with average pooling.

The generated stereo volumes ($F_{sv}^i, i=1,2,3$) are in camera frustum space and exist the issue of object distortion. To solve the issue, we introduce a multi-scale BEV projection and fusion module to convert camera frustum space to 3D world space and then generate 3D geometry-aware features.

\subsubsection{Multi-Scale BEV Projection and Fusion (MPF)} 
\label{sec:mpf}
As shown in Fig. \ref{fig:arch}(c), the MPF module first  transforms multi-scale stereo volumes ($F_{sv}^i, i=1,2,3$) in camera frustum space to multiple geometry volumes ($F_{gv}^i, i=1,2,3$) in 3D world space. After that, the MPF module converts geometry volumes in 3D space to the features in bird's eye view (BEV) and performs a multi-level fusion to generate geometry-aware features ($F_{gf}^i, i=1,2,3$) for 3D prediction.

To transform stereo volume in camera frustum space to geometry volume in 3D space, we adopt volume transformation operation introduced in DSGN \cite{chen2020dsgn}. Specifically, we first generate a regular voxel gird in 3D space and project each voxel in grid into camera frustum space with camera internal parameters. After that, we perform a reversing 3D projection to project the corresponding feature in stereo volume to that in  geometry volume.   With multi-scale stereo volumes ($F_{sv}^i, i=1,2,3$), we adopt the voxel gird with same size to generate multiple  geometry volumes ($F_{gv}^i, i=1,2,3$) with the same resolution. After that,  we convert the geometry volumes to the BEV features  ($F_{bev}^i  \in \mathbb{R}^{(C*Y) \times X \times Z}, i=1,2,3$) by flattening geometry volumes along channel and $y$ dimensions.

With the BEV features ($F_{bev}^i, i=1,2,3$), we perform a multi-level fusion to generate the enhanced geometry-aware features ($F_{gf}^i \in \mathbb{R}^{C' \times X \times Z}, i=1,2,3$). Specifically, We first use a 2D convolution to generate geometry-aware feature $F_{gf}^1$. At the same time, We  concatenate $F_{gf}^1$ with $F_{bev}^2$ and fed it to a 2D convolution to generate geometry-aware feature $F_{gf}^2$. Similarly, we generate geometry-ware feature $F_{gf}^3$. We also map semantic features $F_l^3$ to 3D space and concatenate it with geometry-aware feature $F_{gf}^3$ like \cite{chen2020dsgn}.
Finally, the geometry-aware feature $F_{gf}^3$ is used to perform 3D prediction.
\begin{figure}[t]
\centering
\includegraphics[width=0.999\linewidth]{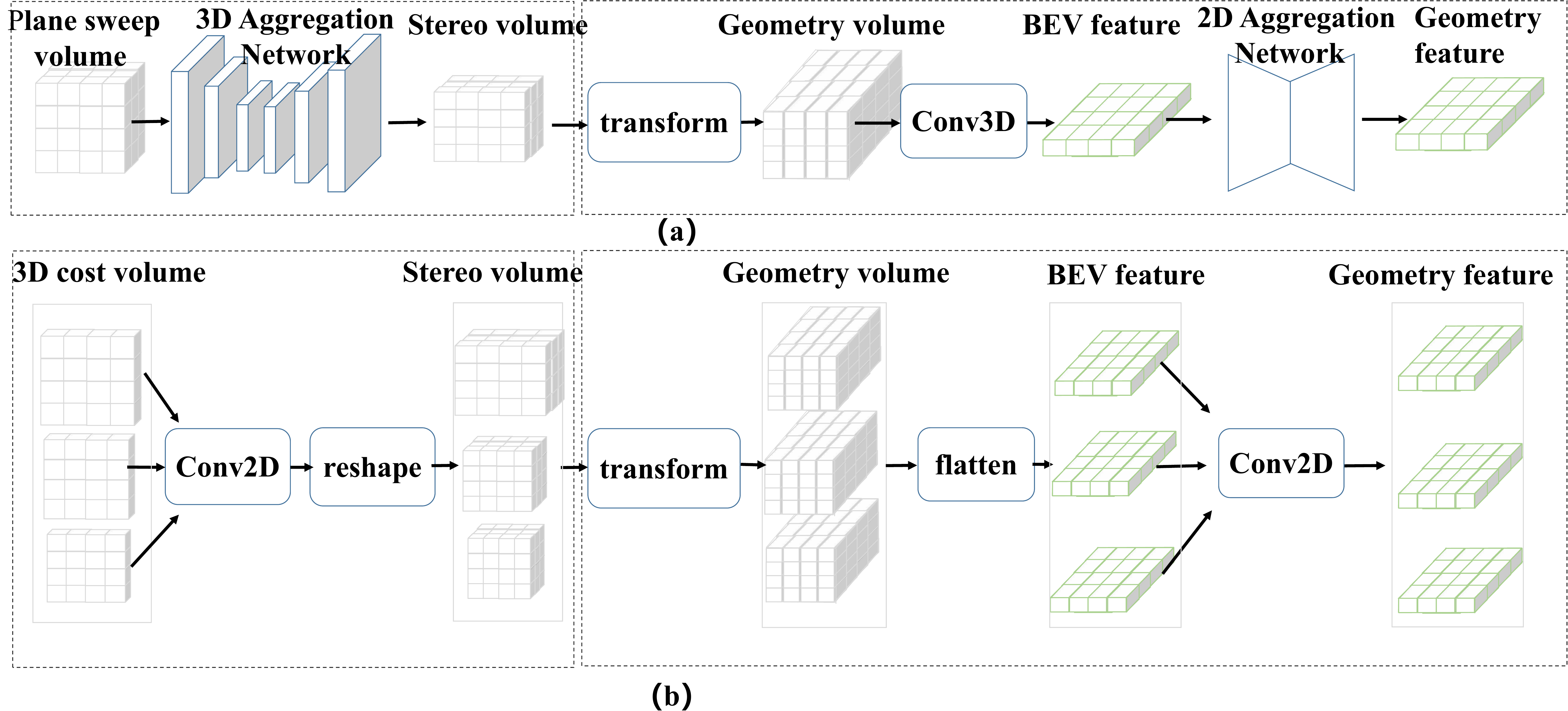}
\caption{A detailed comparison between DSGN (a) and our EGFG (b). Our EGFG adopts deep multi-scale information fusion for stereo volume generation (left part) and geometry-aware feature generation (right part) which avoids complex 3D and 2D aggregation network in DSGN \cite{chen2020dsgn}.}
\label{fig:detailcomp}
\end{figure}

\textbf{Difference to DSGN} Fig. \ref{fig:detailcomp} gives a comparison between DSGN \cite{chen2020dsgn} and our EGFG. Both DSGN and our EGFG extract stereo volume in camera frustum space and convert camera frustum space to 3D and BEV spaces for geometry-aware feature generation. Though they adopt a similar pipeline, they have  significant differences as follows: (1) The goal is different. DSGN aims to explore an accurate 3D detector without considering computational costs, while our EGFG aims to explore a fast stereo 3D object detector with a best trade-off between speed and accuracy,  (2) The key idea of geometry-aware feature generation is different. DSGN employs the heavy 3D and 2D aggregation networks to extract discriminative geometry-aware features, while our EGFG adopts deep multi-scale fusion to generate discriminative geometry-aware features. (3) We observe that it achieves a poor performance without the heavy 3D and 2D aggregation networks used in DSGN \cite{chen2020dsgn} (see Table 4).

\begin{table*}[t]
\caption{Comparison ($AP_{3D}$) of some state-of-the-art 3D stereo object (car) detection methods on both KITTI validation and test sets. The inference time, except YOLOStereo3D, is taken from the leaderboards on official KITTI website.  YOLOStereo3D and our ESGN are reported on NIVIDA RTX 3090.}
\label{tab:ap3d}
\scriptsize
\begin{center}
\setlength{\tabcolsep}{4mm}
\resizebox{\linewidth}{!}
{\begin{tabular}{l|c|c|c|c|c|c|c}
\hline
\multirow{2}*{Method} & \multirow{2}*{Time} &\multicolumn{3}{c|}{\textbf{mAP (test)}} & \multicolumn{3}{c}{IoU $ >$ 0.7(validation)} \\ \cline{3-8}
 &    &\textbf{Moderate}& Easy  & Hard  & \textbf{Moderate}& Easy &        Hard \\ \hline\hline
 
TL-Net \cite{TL-net}& - & 4.37 & 7.64  & 3.74& 14.26  & 18.15 & 13.72  \\
Stereo RCNN \cite{Li_StereoRCNN_CVPR_2019}  & 300ms & 30.23 & 47.58  & 23.72 & 36.69 & 54.11  & 31.07 \\ 
IDA3D \cite{Peng_IDA3D_CVPR_2020}&300ms &29.32&45.09  &23.13 &37.45 &54.97 &32.23\\
PL: F-PointNet \cite{wang_PseudoLiDAR_2019} &400ms& 26.70 & 39.70  & 22.30 &39.8 & 59.4 & 33.5  \\ 
PL: AVOD \cite{wang_PseudoLiDAR_2019}  & 400ms& 34.05 & 54.53  & 28.25  & 45.3 & 61.9  & 39   \\ 
PL++: AVOD \cite{you_Pseudo-LiDAR++_2020}  & 400ms & -&-&-& 46.8&63.2   & 39.8 \\ 
PL++: P-RCNN \cite{you_Pseudo-LiDAR++_2020}  & 400ms  & 42.43& 61.11 & 36.99& 44.9& 62.3  & 41.6 \\ 
OC-Stereo \cite{Pon_ocstereo_2020} & 350ms & 37.60 & 55.15  & 30.25 & 48.34 & 64.07  &40.39 \\ 
ZoomNet \cite{xu_Zoomnet_2020} & 300ms  & 38.64& 55.98 &  30.97& 50.47& 62.96  & 43.63  \\ 
Disp R-CNN \cite{sun_disprcnn_2020} & 387ms & 45.78 & 68.21  &37.73& 47.73 & 64.29  & 40.11   \\
DSGN \cite{chen2020dsgn}&  670ms  & 52.18&73.50&45.14& 54.27 &72.31& 47.71\\
CG-Stereo \cite{Li_CGstereo_2020} &570ms &53.58&74.39&46.50 & 57.82 &76.17&54.63\\
LIGA\cite{guo2021liga}&400ms &64.66&81.39&57.22& 67.06& 84.92 &63.80\\
\hline
RT3DStereo \cite{RT3DStereo}& 80ms &23.28 & 29.90 & 18.96&-&-&- \\
Stereo-Centernet \cite{shi2021sc}&40ms&31.30&49.94&25.62&41.44&55.25&35.13\\
RTS3D \cite{Li_RTS3D_AAAI_2021}  & 39ms& 37.38 & 58.51  & 31.12&44.5& 63.65&37.48  \\
RT3D-GMP \cite{RT3D-GMP}&60ms&38.76&45.79&30.00&-&-&-\\
YOLOStereo3D \cite{liu2021yolostereo3d} & 50ms  &  41.25& 65.68 &  30.42& 46.58& 72.06 & 35.53 \\
\textbf{ESGN (Ours)}  & 62ms &\textbf{46.39}& \textbf{65.80}&\textbf{38.42}& \textbf{52.33}&\textbf{72.44}& \textbf{43.74} \\ \hline  
\end{tabular}}
\end{center}
\end{table*}

\subsection{Deep Geometry-Aware Feature Distillation (DGFD)}
\label{sec:dgfd}
LiDAR-based 3D object detection has a higher accuracy than stereo 3D object detection. 
To bring this gap, Guo \textit{et al.} \cite{guo2021liga} proposed a novel feature distillation approach (LIGA) for stereo 3d object detection. LIGA  attaches a single-level feature distillation on the output geometry-aware features of deep stereo geometry network (DSGN) \cite{chen2020dsgn}. We argue that this single-level distillation can not provide a deep supervision. To this end, we propose a deep  geometry-aware feature distillation (DGFD) scheme, where a LiDAR-based 3D detector is designed to generate multi-level LiDAR features and then provide deep multi-level supervision for stereo feature learning.

\begin{figure}[t]
\centering
\includegraphics[width=0.999\linewidth]{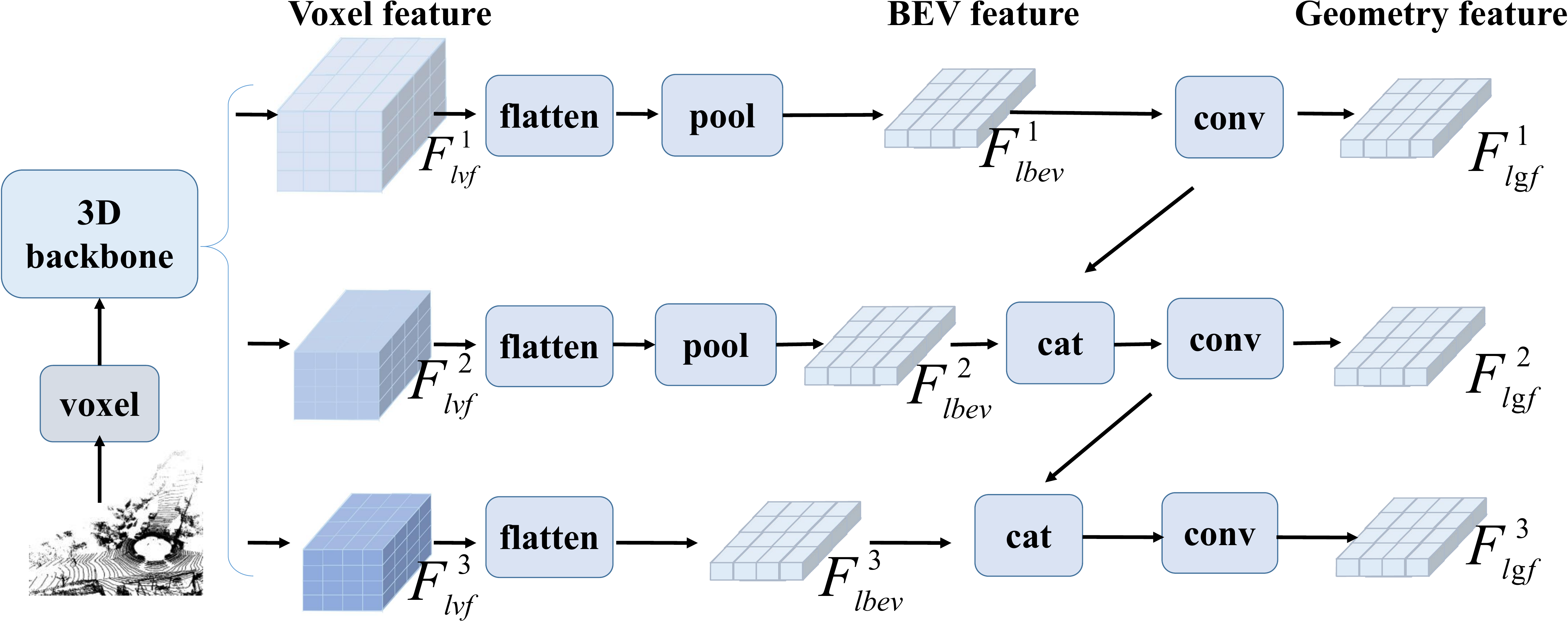}
\caption{Detailed architecture of our designed LiDAR detector for distillation. The LiDAR detector first generates multi-scale voxel feature maps by the voxel operation and a sparse 3D convolution backbone like \cite{SECONDyan}. Then, these multi-scale voxel features are used to generate multiple geometry features.}
\label{fig:fused}
\end{figure}

Fig.~\ref{fig:fused} shows the architecture of our designed LiDAR 3D detector. Given the LiDAR point cloud, we convert raw point cloud representation into  a voxel representation and  use a spare 3D 
convolutional backbone \cite{SECONDyan} to extract multi-scale LiDAR voxel features ($F_{lvf}^i, i=1,2,3$). Then, we  flatten  LiDAR voxel features along channel and $y$ dimensions and pool the BEV features to the same solution by using average pooling. We call the resized output features as LiDAR BEV features ($F_{lbev}^i, i=1,2,3$). After that, we perform a multi-level fusion to output LiDAR geometry-aware features ($F_{lgf}^i, i=1,2,3$). Finally, we perform LiDAR 3D prediction on $F_{lgf}^3$.

We first train this LiDAR 3D detector on point cloud data. After that, the LiDAR geometry-aware features ($F_{lgf}^i \in \mathbb{R}^{C'\times X\times Z}, i=1,2,3$) are used to guide stereo geometry-aware feature learning. Specifically, we minimize the feature difference $L_{dif}$ between LiDAR geometry-aware features and stereo geometry-aware features at mutliple levels as follows:
\begin{equation}
\begin{array}{c}
L_{dif}  =  \displaystyle\sum_{i=1,2,3}  \frac{1}{N} \left | M_{fg}M_{sp}(g(F_{gf}^i) - F_{lgf}^i  )\right |^2 
\end{array}
\end{equation}
where $i$ represents the scale index, $g$ represents a single $1\times1$ convolution, $M_{fg}$ is the foreground mask, $M_{sp}$ is LiDAR sparse mask, and $N$ is the number of sparse foreground mask.

\subsection{Prediction Heads}
\label{sec:head}

{\bf 3D Detection Head} Similar to \cite{chen2020dsgn,guo2021liga}, we perform 3D detection by a classification head and a regression head. During the training, the total loss of 3D detection can be written as $L_{3d}=L_{cls}+L_{l1}+L_{dir}+L_{iou}$, where $L_{cls}, L_{l1}, L_{iou}$ respectively represent classification loss,
box regression loss, and direction classification loss in \cite{guo2021liga,SECONDyan}, and $L_{iou}$ represents the rotated IoU loss \cite{ZhouIOULOSS}.

{\bf Auxiliary Heads} Similar to \cite{chen2020dsgn,guo2021liga}, we add two auxiliary heads during training, including a deep estimation head and a 2D detection head. The depth estimation head consists of several 
convolutions and is attached on stereo volume $F_{sv}^3$ in Fig. \ref{fig:arch}(b). The ground-truth of depth estimation is transformed to form point cloud data. The 2D  detection head is attached on the backbone feature $F_{l}^1$ in Fig. \ref{fig:arch}(a). During the training, the auxiliary loss can be written as $L_{aux}=L_{depth}+L_{2d}$, where $L_{depth}$ represents depth estimation loss and $L_{2d}$ represents 2D object detection loss.

\begin{table*}[t]
\caption{Comparison ($AP_{BEV}$) of state-of-the-art 3D object (car) detection methods on KITTI validation set. The inference time, except YOLOStereo3D, is from the  official KITTI leaderboards. YOLOStereo3D and our ESGN are reported on NIVIDA RTX 3090.}
\label{tab:bev}
\scriptsize
\begin{center}
\setlength{\tabcolsep}{4mm}
\resizebox{\linewidth}{!}{
\begin{tabular}{l|c|c|c|c|c|c|c}
\hline
\multirow{2}*{Method} & \multirow{2}*{Time} & \multicolumn{3}{c|}{IoU $>$ 0.7}& \multicolumn{3}{c}{IoU $ >$ 0.5} \\ \cline{3-8}
  &  & \textbf{Moderate}& Easy  & Hard & \textbf{Moderate}& Easy & Hard \\ \hline\hline
TL-Net \cite{TL-net} &-    &	21.88   &29.22	 &18.83   &45.99&62.46&41.92  \\ 
Stereo RCNN \cite{Li_StereoRCNN_CVPR_2019} & 300ms &48.30&68.50		&	41.47 &74.11& 87.13&58.93		\\ 
IDA3D \cite{Peng_IDA3D_CVPR_2020}&300ms& 50.21& 70.68 &42.93&76.69&88.05& 67.29\\

PL: F-PointNet \cite{wang_PseudoLiDAR_2019} & 400ms &51.8&72.8&44       &77.6	&89.8	&68.2
\\ 
PL: AVOD \cite{wang_PseudoLiDAR_2019}& 510ms &39.2&60.7&37&65.1&76.8&56.6	
\\ 
PL++: AVOD \cite{you_Pseudo-LiDAR++_2020}&400ms&56.8&74.9	&49&77.5&89&68.7
\\ 
PL++: PIXOR \cite{you_Pseudo-LiDAR++_2020}& 400ms &	61.1& 79.7		&	54.5	&75.2&89.9&67.3
\\   
PL++: P-RCNN \cite{you_Pseudo-LiDAR++_2020} & 400ms &	56&73.4&52.7 &76.6&88.4&69		
\\ 
OC-Stereo \cite{Pon_ocstereo_2020} &350ms&	65.95& 77.66&51.20&80.63&90.01&71.06		
\\ 
ZoomNet \cite{xu_Zoomnet_2020} &300ms&66.19&78.68&57.60 &88.40&90.62		&71.44	
\\ 
Disp R-CNN \cite{sun_disprcnn_2020} & 387ms &	64.38&77.63&	50.68 &80.45&90.67&71.03\\
DSGN \cite{chen2020dsgn}&  670ms & 63.91&83.24&57.83& - & -& - \\
CG-Stereo \cite{Li_CGstereo_2020}&570ms & 68.69&87.31&65.80& 88.58& 97.04 & 80.34\\
LIGA \cite{guo2021liga}&400ms&77.26&89.35&69.05& 90.27& 97.22 & 88.36 \\

\hline
Stereo-Centernet \cite{shi2021sc}&40ms& 53.27 &71.26 &45.53&-&-&- \\
RTS3D \cite{Li_RTS3D_AAAI_2021} & 39ms & 56.46& 76.56& 48.20& 78.70& 90.41&70.03\\ 
YOLOStereo3D \cite{liu2021yolostereo3d} & 50ms  &  55.22& 80.69 &43.47&79.62& 96.52& 62.50\\
\textbf{ESGN (Ours)}  & 62ms &\textbf{63.86}& \textbf{82.29}&\textbf{54.63}&\textbf{82.22}&93.05&\textbf{72.25}
\\ \hline
\end{tabular}
}

\end{center}
\end{table*}

\section{Experiments}
\subsection{Dataset and Implementation Details}

\textbf{Dataset} We perform the experiments on the classical KITTI dataset \cite{kitti}. The KITTI dataset consists of 7,481 training paired images and 7,518 test paired images. In addition, the dataset provides the LiDAR point cloud data for each RGB image.
Following the existing works \cite{Li_StereoRCNN_CVPR_2019,wang_PseudoLiDAR_2019,Pon_ocstereo_2020}, we spilt the original training images into the training set and the validation set. The training set has 3,712 paired images and the validation set has 3,769 paired images. 
For ablation study, we train our proposed ESGN on the training set with 3,712 images and evaluate it on the validation set.
For state-of-the-art comparison, we train our proposed ESGN  on the original training images with 7,481 images and submit the results on the test set  to the official evaluation server for performance evaluation.

\textbf{Implementation Details} We implement our proposed ESGN on a single NIVIDA RTX3090 GPU. To generate LiDAR features for deep distillation,  we first train our designed LiDAR-based detector on the training set using LiDAR point cloud data. We adopt Adam  for optimization and set the batch size as 2. There are 80 epochs, where the learning rate is set as 0.003 and decreases at epoch 35 and 45. After that, we train our ESGN on the training set using stereo images. We adopt Adam  for optimization and set the batch size as 1. There are 55 epochs, where the learning rate is set as 0.001 and decreases at epoch 50 by a factor of 10.

For KITTI dataset, we set the detection region in range [-30, 30] $\times$ [-1, 3] $\times$ [2, 59.6] (meters). The voxel size in stereo is set as [0.4m, 0.8m, 0.4m] for regular voxel grid generation (see Sec. \ref{sec:scr}), while  the voxel size in LiDAR is set as [0.05m, 0.1m, 0.05m] for LiDAR voxel representation.

\subsection{Comparison With State-Of-The-Art Methods}
We first compare our proposed ESGN with some state-of-the-art methods on both KITTI  test and validation sets. 
According to the degree of occlusion and truncation, the validation and test sets are respectively divided into three subsets: \texttt{easy}, \texttt{moderate} and \texttt{hard}. 
Tab. \ref{tab:ap3d} gives the state-of-the-art  comparisons in terms of speed and $AP_{3d}$.  Compared to the high-accuracy DSGN \cite{chen2020dsgn} and LIGA \cite{guo2021liga}, our proposed ESGN is 11.2 and 6.5 times faster in speed. Among these state-of-the-art methods, 
RT3DStereo \cite{RT3DStereo}, Stereo-Centernet \cite{shi2021sc}, RTS3D \cite{Li_RTS3D_AAAI_2021}, RT3D-GMP\cite{RT3D-GMP}, and YOLOStereo3D \cite{liu2021yolostereo3d} belong to fast stereo 3D objection detection approaches, which all have the speed of less than 100$ms$. For example, YOLOStereo3D \cite{liu2021yolostereo3d} achieves an $AP_{3d}$ of 41.25\% on moderate test set at the speed of 50$ms$. Compared to these fast stereo 3D detection approaches, our proposed ESGN achieves the best  accuracy  on all three subsets of both test and validation sets. For example,  our proposed ESGN outperforms YOLOStereo3D by an absolute gain of 5.14\% on KITTI moderate test set at a comparable speed.

Tab. \ref{tab:bev} further provides the state-of-the-art comparison in terms of both speed and $AP_{bev}$ on KITTI validation set. We show the reasults under two evaluation metrics (\textit{i.e.,} IoU $>$ 0.5 and IoU $>$ 0.7). On the moderate set, our ESGN outperforms these fast stereo object detection approaches under these two evaluation metrics. Moreover, we observe that our ESGN is  much better under the stricter evaluation metric (IoU $>$ 0.7).  For example,  our proposed ESGN outperforms YOLOStereo3D by an absolute gain of 8.64\% on moderate set with the evaluation metric of IoU $>$ 0.7.

\begin{table}[t]
\begin{center}
\caption{Impact of integrating EGFG (Sec. \ref{sec:egfg}) and DGFD (Sec. \ref{sec:dgfd}) modules into the baseline on KITTI validation set.}
\label{tab:twomodules}
\begin{tabular}{ccc|c|c|c}
\hline
Baseline & EGFG & DGFD   & \textbf{Moderate}  & Easy & Hard \\ \hline\hline
\checkmark   &  &    & 15.54 & 23.89 & 13.32 \\
\checkmark   & \checkmark &    & 49.69 & 68.05 & 41.40 \\
\checkmark   & \checkmark & \checkmark   & 52.33 & 72.44 & 43.74 \\
\hline
\end{tabular}
\end{center}
\end{table}

\begin{figure*}[t]
\centering
\includegraphics[width=0.99999\linewidth]{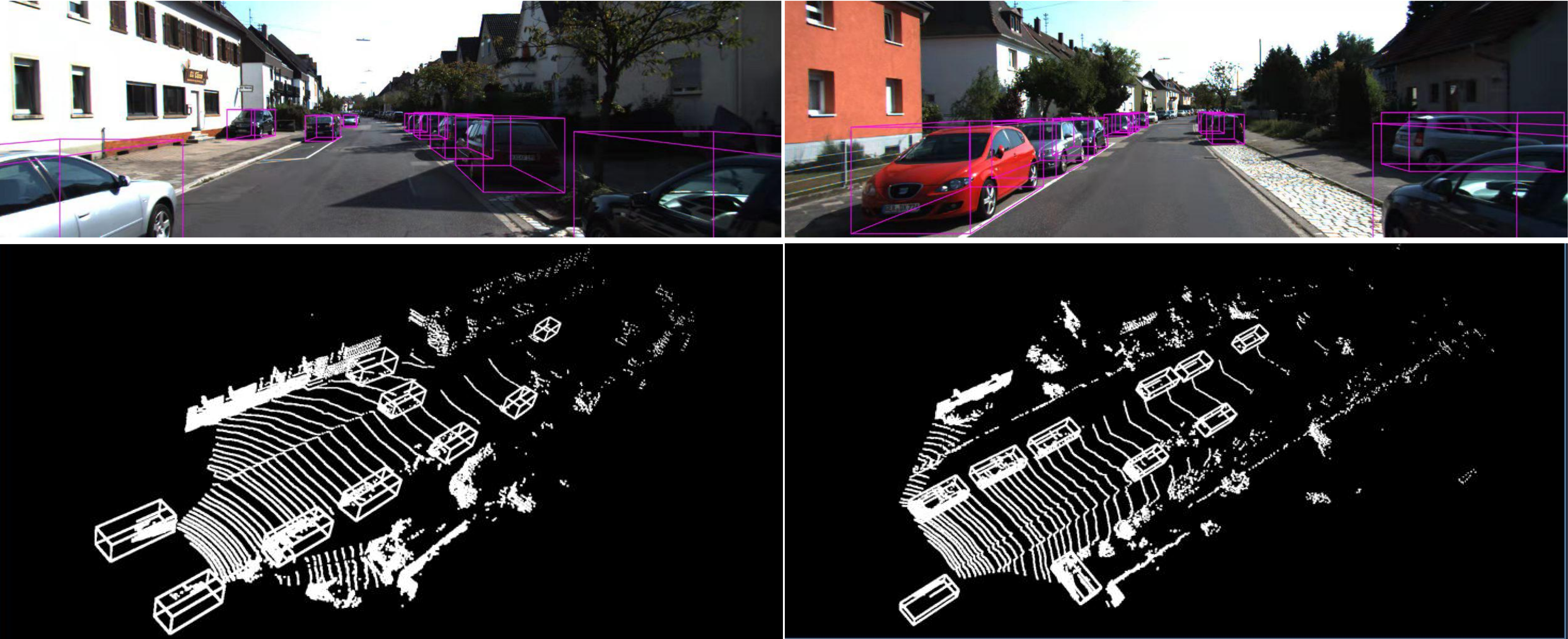}
\caption{Qualitative stereo 3D results of our ESGN. The top row shows 3D detection results in image viewpoint, and the bottom row shows 3D detection results in 3D space. Our ESGN detects both large and small objects accurately.}
\label{fig:results}
\end{figure*}

\subsection{Ablation Study}
In this subsection, we conduct the ablation study to demonstrate the effectiveness of different modules in our proposed ESGN. All the results in this subsection are evaluated on KITTI validation set under the evaluation metric of IoU $>$ 0.7.

We first show the impact of progressively integrating different modules, including EGFG in Sec. \ref{sec:egfg} and DGFD in Sec. \ref{sec:dgfd}, into the baseline.   Tab. \ref{tab:twomodules} shows the results on three subsets. Our baseline directly adopts the single stereo volume $F_{sv}^1$ for stereo 3D detection using 3D head like method \cite{liu2021yolostereo3d}. On moderate subset, our baseline achieves a very low $AP_{3d}$  of 15.54\% on moderate subset, due to very simple and plain design. Then, we integrate our EGFG into the baseline, where EGFG fuses deep multi-level information for both stereo volume and geometry-aware feature generations. Our EGFG achieves an $AP_{3d}$  of 49.69\%, which demonstrates the effectiveness of our EGFG module. Finally, we further integrate our DGFD into them, which achieves an $AP_{3d}$  of 52.33\%. It demonstrates that deep distillation is useful for 3D detection.

\begin{table}[t]
\begin{center}
\caption{Impact of two modules, including SCR (Sec. \ref{sec:scr}) and MPF (Sec. \ref{sec:mpf}), in our EGFG module (Sec. \ref{sec:egfg}) on KITTI validation set. `single' represents using only one of three scales without multi-scale fusion, while `Multi.' represents using three scales with multi-scale fusion.}
\label{tab:EGFG}
\begin{tabular}{cc|cc|c|c|c}
\hline
\multicolumn{2}{c|}{SCR}  & \multicolumn{2}{c|}{MPF} & \multirow{2}*{\textbf{Mod}}   & \multirow{2}*{Easy}  & \multirow{2}*{Hard} \\ \cline{1-4}
Sing.  &  Multi.  &  Sing. & Multi. &   &   &   \\ \hline
\hline
\checkmark   &  &  &    & 15.54 & 23.89 & 13.32 \\
\checkmark   &  & \checkmark &   & 37.87  & 57.12& 31.52 \\
   & \checkmark & \checkmark &    & 46.24 & 64.98 & 38.79 \\
   &\checkmark   &  & \checkmark   & 49.69 & 68.05 & 41.40 \\
\hline
\end{tabular}
\end{center}
\end{table}

\begin{table}[t]
\begin{center}

\caption{Impact of single-level distillation and deep distillation in our DGFD module (Sec. \ref{sec:dgfd}) on KITTI validation set.}
\label{tab:DGFD}
\begin{tabular}{l|c|c|c}
\hline
Method & \textbf{Moderate} & Easy  & Hard \\ \hline
\hline
No distillation & 49.69 & 68.05  & 41.40 \\
Single-level distillation   & 51.40 & 72.25  & 43.31 \\
Deep distillation & 52.33 &   72.44  & 43.74 \\
\hline
\end{tabular}
\end{center}
\end{table}

We further show the impact of different modules in our EGFG in Tab. \ref{tab:EGFG}. Our EGFG contains a stereo correlation and reprojection (SCR) module and a multi-scale BEV projection and fusion (MPF) module. When using only single scale SCR (baseline in Tab. \ref{tab:twomodules}), it achieves an $AP_{3d}$ of 15.54\% on moderate subset. When using single-scale SCR and single-scale MPF, it achieves an  $AP_{3d}$ of 37.87\%. Note that, this single-scale SCR and single-scale MPF setting is similar to the light-weight DSGN in which the heavy 3D and 2D aggregation networks are removed. The light-weight DSGN is inferior to our ESGN.  When using multi-scale SCR and single-scale MPF, it achieves an $AP_{3d}$ of 46.24\%, which provides 8.37\% improvement. Namely, single-scale SCR can not extract multi-scale geometry information for stereo volume generation. Based on multi-scale SCR, we perform multi-scale MPF and further improve the performance by 3.45\%. Namely, multi-scale BEV projection and fusion further enhances geometry-aware features. 

We also compare single-level distillation and deep distillation in Tab. \ref{tab:DGFD}. Single-level distillation only guides feature learning at single geometry-aware feature $F_{gf}^3$ like \cite{guo2021liga}, while deep distillation guides  feature learning at multiple geometry-aware features $F_{gf}^i, i=1,2,3$. Compared to no distillation design, these two distillation strategies respectively provide 1.71\% and 2.64\% improvements on moderate set. Compared to single-level distillation, our proposed deep distillation has 0.93\% improvement. It demonstrates that deep multi-level distillation can better guide geometry-aware feature learning in multiple levels. In addition, our deep distillation does not add extra computation costs during inference. 

Finally, we provide some qualitative 3D detection results of our proposed ESGN in Fig. \ref{fig:results}. The results in image viewpoint (top) and corresponding 3D space (bottom) are both provided. We observe that our ESGN can detect both large and small objects accurately.
\section{Conclusion}
In this paper, we have proposed an efficient stereo geometry network (ESGN) for fast 3D object detection. The key module is a novel efficient geometry-aware feature generation (EGFG) module that first generates multi-scale stereo volumes by a stereo correlation and reprojection (SCR) module and second generates geometry-aware features by a multi-scale BEV projection and fusion (MPF) module. With deep multi-scale fusion, our EGFG module generates discriminative  geometry-aware features without heavy aggregation operations. We also introduce a deep geometry-aware feature distillation scheme to guide feature learning with a LiDAR-based 3D detector. We perform experiments on KITTI dataset. Our ESGN achieves a best trade-off between speed and accuracy. Compared to the high-accuracy DSGN, our ESGN is 11.2 times faster in speed. Compared to the fast YOLOStereo3D, our ESGN achieves an $AP_{3d}$ improvement of 5.14\%  at comparable fast speed. We hope that our ESGN can provide more possible ways for fast 3D object detection.



\end{document}